\documentclass[runningheads]{llncs}

\usepackage{eccv}

\usepackage{eccvabbrv}

\usepackage{graphicx}
\usepackage{booktabs}

\usepackage[accsupp]{axessibility}  

\usepackage{hyperref}

\usepackage{orcidlink}

\usepackage{url}            
\usepackage{booktabs}       
\usepackage{amsfonts}       
\usepackage{nicefrac}       
\usepackage{microtype}      
\usepackage{lipsum}
\usepackage{fancyhdr}       
\usepackage{graphicx}       
\usepackage{multirow}
\usepackage{booktabs}
\usepackage{amsmath}
\usepackage{bm}
\usepackage{amssymb}
\usepackage{comment}
\usepackage{placeins}
\usepackage{algorithm}
\usepackage{algorithmic}
\usepackage[accsupp]{axessibility}
\usepackage{multicol}
\usepackage{wrapfig}
\usepackage{subcaption} 

\usepackage{xcolor}
\usepackage{pifont}

\newcommand{\cmark}{\textcolor{green!60!black}{\ding{51}}} 
\newcommand{\xmark}{\textcolor{red}{\ding{55}}} 

\begin{document}

\title{WildRelight: A Real-World Benchmark and Physics-Guided Adaptation for Single-Image Relighting} 

\titlerunning{WildRelight}

\author{Lezhong~Wang \and Mehmet~Onurcan~Kaya  
\and Siavash~Bigdeli \and Jeppe~Revall~Frisvad}

\authorrunning{LZ.~Wang et al.}

\institute{Technical University of Denmark, Kongens Lyngby, Denmark \\
\email{\{lewa, monka, sarbi, jerf\}@dtu.dk, lezhong.wang@inria.fr}
}
\maketitle

\begin{abstract}
Recent single-image relighting methods, powered by advanced generative models, have achieved impressive photorealism on synthetic benchmarks. However, their effectiveness in the complex visual landscape of the real world remains largely unverified. A critical gap exists, as current datasets are typically designed for multi-view reconstruction and fail to address the unique challenges of single-image relighting. 
To bridge this synthetic-to-real gap, we introduce \textbf{WildRelight}, the first in-the-wild dataset specifically created for evaluating single-image relighting models. WildRelight features a diverse collection of high-resolution outdoor scenes, captured under strictly aligned, temporally varying natural illuminations, each paired with a high-dynamic-range environment map. 
Using this data, we establish a rigorous benchmark revealing that state-of-the-art models trained on synthetic data suffer from severe domain shifts. 
The strictly aligned temporal structure of WildRelight enables a new paradigm for domain adaptation. We demonstrate this by introducing a physics-guided inference framework that leverages the captured natural light evolution as a self-supervised constraint. By integrating Diffusion Posterior Sampling (DPS) with temporal Sampling-Aware Test-Time Adaptation (TTA), we show that the dataset allows synthetic models to align with real-world statistics on-the-fly, transforming the intractable sim-to-real challenge into a tractable self-supervised task. 
The dataset and code will be made publicly available to foster robust, physically-grounded relighting research.
\end{abstract}    
\begin{figure}
    \centering
    \includegraphics[width=1.0\textwidth]{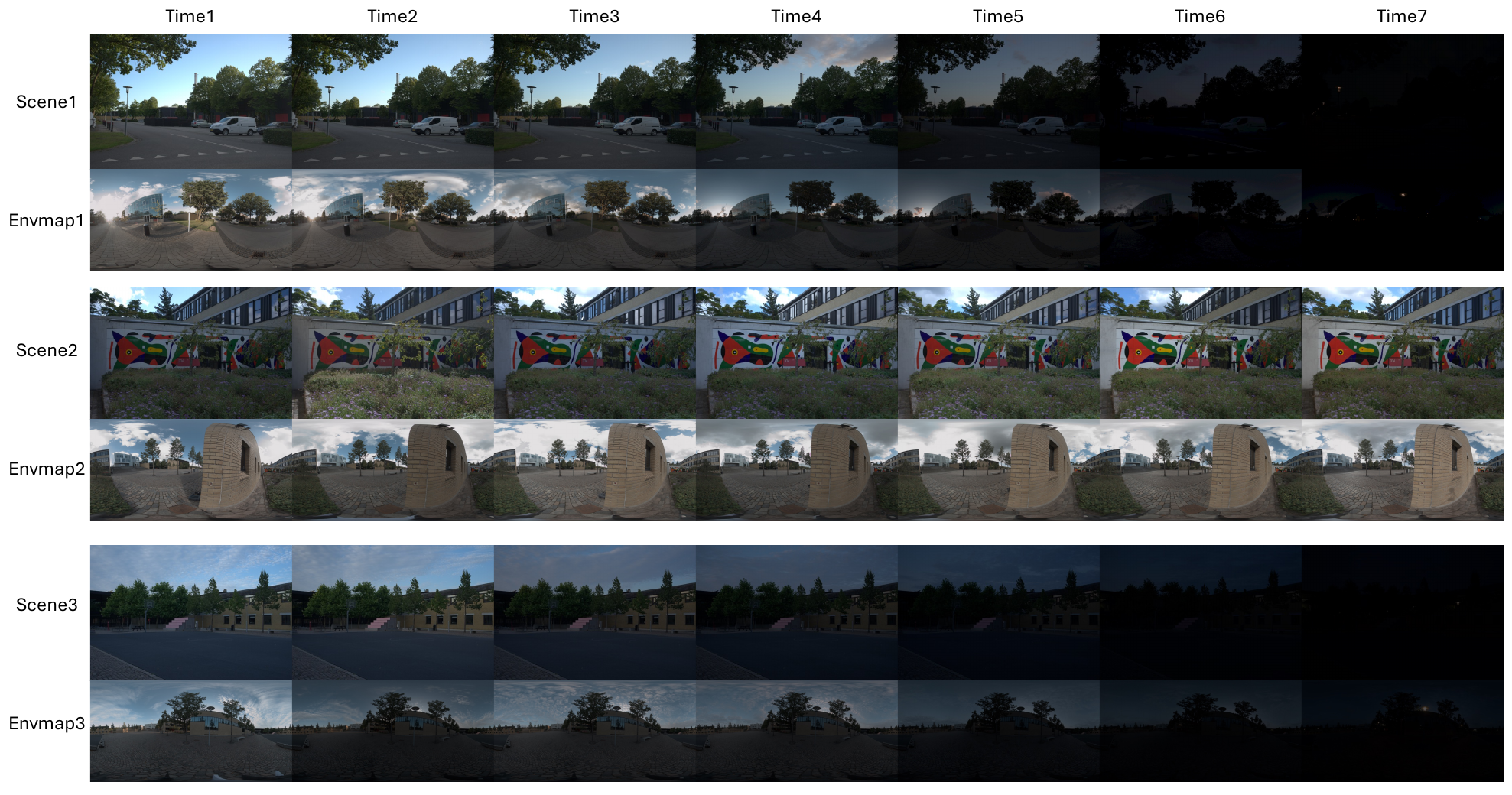} 
    \captionof{figure}{Example image and illumination pairs from our \textit{WildRelight} dataset  %
    }
    \label{fig:teaser}
\end{figure}

\section{Introduction}

Manipulating the illumination within a single photograph is a long standing goal in computer vision and graphics, with profound applications in computational photography, filmmaking, and augmented reality~\cite{debevec2023modeling, barron2014shape,toschi2023relight,lombardi2015reflectance,li2021openrooms}. Recently, the field has seen a dramatic leap forward, propelled by the expressive power of deep generative models~\cite{liu2023openillumination,luo2024intrinsicdiffusion,liang2025diffusion}. State of the art methods can now decompose a single image into its intrinsic components (albedo, geometry, illumination) and rerender the scene under novel lighting with stunning photorealism~\cite{haque2023instruct,luo2024intrinsicdiffusion,zeng2024rgb}.

Despite this remarkable progress, a critical question remains unanswered: how well do these models perform outside the sanitized confines of synthetic data? The training and, more importantly, the quantitative evaluation of most inverse rendering models are predominantly conducted on synthetic datasets~\cite{li2022phyir, zhang2021nerfactor,liang2025diffusion,zeng2024rgb}. While invaluable for development, synthetic data often fails to capture the intricate complexities of real-world light transport, such as subtle atmospheric scattering, complex indirect illumination, and the rich, non-ideal material properties of natural surfaces. This creates a significant domain gap, where a model's impressive performance on synthetic benchmarks may not translate to practical, real-world applications.

To address this critical need, we introduce \textbf{WildRelight}. Unlike large-scale pretraining corpora, WildRelight is designed as a high-precision evaluation benchmark (akin to Middlebury\cite{scharstein2003high} or DTU \cite{jensen2014large}). We prioritize strict pixel-alignment and radiometric accuracy over quantity, as misalignment renders large-scale data useless for physical validation. 
We systematically captured a diverse array of outdoor scenes at various times of the day, from the golden hour to the harsh midday sun, to encompass a wide spectrum of natural illumination. Our core technical contribution lies in a precise data acquisition pipeline. For each high resolution image captured with a primary camera, we simultaneously recorded a full 360° High Dynamic Range (HDR) environment map with a panoramic camera. We utilized a custom built rig to precisely colocate the optical centers of both cameras. This strict spatial alignment ensures the captured HDR environment map serves as an accurate ground truth representation of the incident illumination for its corresponding single-view image.

To demonstrate the distinct research opportunities unlocked by this protocol, we present a reference application: a unified framework comprising Physics-Guided Inverse Rendering and Sampling-Aware Test-Time Adaptation. This methodology is explicitly designed to validate the utility of our captured natural illumination evolution. By leveraging the strict temporal alignment provided by WildRelight, we show how the complex synthetic-to-real domain adaptation challenge can be reformulated into a tractable, real-world self-supervised task. This application serves as a proof-of-concept, illustrating how the dataset empowers models to bridge the domain gap through instance-specific adaptation using only the available test-time observations.

We provide a detailed comparison of WildRelight against existing real-world multi-illumination datasets in \autoref{tab:dataset_comparison}.
These datasets can be broadly categorized. 
First, controlled laboratory datasets, such as OpenIllumination \cite{liu2023openillumination} and ReNe \cite{toschi2023relight}, use light stages or robotics to capture high fidelity object data, often with precise One-Light-at-a-Time (OLAT) sources and perfect viewpoint alignment. While invaluable for BRDF estimation, they lack the geometric complexity and rich, full spectrum illumination of ``in-the-wild'' scenes.
Second, multi-view ``in-the-wild'' datasets, like NeRF-OSR \cite{rudnev2022nerf} and Objects With Lighting \cite{Ummenhofer2024OWL}, capture real natural lighting but are designed for multi-view 3D reconstruction. Their primary data consists of moving camera trajectories, meaning they lack the static, cross-illumination viewpoint alignment necessary for evaluating single-image relighting.
Finally, other single-view datasets, such as Murmann et al. \cite{murmann2019dataset} or LSMI \cite{aksoy2018dataset}, are typically captured indoors with simple spotlights, lack HDR, do not provide GT environment maps, and often lack strict viewpoint alignment.
In contrast, WildRelight fills a critical, unaddressed gap by being the first dataset to combine all the necessary features for our task: (1) a single-view, static camera setup ensuring strict viewpoint alignment, (2) diverse, ``in-the-wild'' outdoor scenes, and (3) high-fidelity, spatially aligned HDR environment maps for every image.

\begin{table*}[t!]
  \centering
  \caption{
    Comparison with existing real-world, multi-illumination datasets. 
    Our dataset is the first to provide a single-view benchmark for in-the-wild outdoor scenes, featuring HDR GT illumination. 
    Critically, it ensures strict Viewpoint Alignment across all captured illuminations, enabling direct, pixel-aligned evaluation of relighting methods.
  }
  \label{tab:dataset_comparison}
  \resizebox{\textwidth}{!}{%
  \begin{tabular}{@{}llllclccc@{}}
    \toprule
    \textbf{Dataset} & \textbf{Type} & \textbf{Illumination Format} & \textbf{Light Source} & \textbf{\# Illuminations} & \textbf{Purpose} & \textbf{HDR} & \textbf{Alignment} & \textbf{\# Scenes} \\
    \midrule
    NeRF-OSR \cite{rudnev2022nerf} & Outdoor & Environment map & Natural & 5 per scene & Multi-view & \xmark & \xmark & 9 \\
    OpenIllumination \cite{liu2023openillumination} & Objects & OLAT + pattern & Light stage & 142+13 & Multi-view & \cmark & \cmark & 64 \\
    Murmann et al. \cite{murmann2019dataset} & Indoor & Chrome ball & Spotlight & 25 per scene & Single-view & \cmark & \xmark & 1000 \\
    Objects With Lighting \cite{Ummenhofer2024OWL} & Objects & Environment map & Natural & 4 per scene & Multi-view & \xmark & \xmark & 8 \\
    LSMI \cite{aksoy2018dataset} & Indoor & Not Available & Spotlight & 1--3 per scene & Single-view & \xmark & \xmark & 2700 \\
    ReNe \cite{toschi2023relight} & Objects & OLAT & Light stage & 40 OLATs & Multi-view & \xmark & \cmark & 20 \\
    \midrule
    \textbf{Ours} & \textbf{Outdoor} & \textbf{Environment map} & \textbf{Natural} & \textbf{5--7 per scene} & \textbf{Single-view} & \textbf{\cmark} & \textbf{\cmark} & \textbf{30} \\
    \bottomrule
  \end{tabular}%
  }
\end{table*}

\section{Related Work}
\label{sec:related_work}

\paragraph{Single-Image Inverse Rendering.}
Inverse rendering, the ill-posed problem of decomposing a single image into its intrinsic components like albedo, geometry, and illumination~\cite{li2021openrooms,li2022physically,zhu2022irisformer,sengupta2019neural,matusik2003data, lombardi2015reflectance, zhang2021physg,yi2023weakly,zhang2022simbar}, has seen a recent resurgence. Driven by advances in deep generative models, particularly latent diffusion models~\cite{liang2025diffusion,zeng2024rgb,luo2024intrinsicdiffusion}, modern methods are demonstrating remarkable, and often photorealistic, capabilities in single-image relighting. These models learn strong priors about the physical world, enabling them to plausibly rerender a scene under novel lighting. However, this progress is hampered by a significant evaluation gap. Lacking a real-world benchmark with ground truth (GT) illumination, these methods are developed and evaluated almost exclusively on synthetic datasets~\cite{zhang2022modeling,li2021openrooms, liu2023nero,liang2025diffusion,zeng2024rgb}. While synthetic data provides perfect GT, it inherently fails to capture the full complexity of real world phenomena such as the subtle interplay of atmospheric scattering, high frequency shadows from complex foliage, and the rich spectral properties of natural materials. This ``sim-to-real'' gap means that a model's performance on synthetic data is a poor predictor of its efficacy ``in-the-wild". Our work is the first to directly address this critical need by providing a real-world benchmark specifically for single-image relighting.

\paragraph{Controlled Laboratory Datasets.}
To acquire GT data from real-world objects, one dominant line of work relies on highly controlled laboratory environments. This ranges from classical photometric stereo setups with sparse, calibrated lights~\cite{shi2016benchmark,liu2023openillumination, pei2025opensubstance,teufel2025humanolat,zhou2025olatverse} to modern light stages. These sophisticated systems, such as OpenIllumination~\cite{liu2023openillumination}, OpenSubstance~\cite{pei2025opensubstance}, and RelightMyNeRF~\cite{toschi2023relight}, utilize hundreds of controllable LEDs, multiple synchronized cameras, or precise robotics to capture an object's response to illumination. They provide extremely high fidelity data, often including precise 3D geometry from scanners and OLAT measurements, which serve as the definitive GT for material BRDF acquisition. The fundamental limitation of these datasets, however, is twofold: (1) Scope: They are constrained to small scale, isolated objects that can fit inside the capture apparatus, making them unsuitable for studying scenes interaction, architecture, or landscapes. (2) Illumination: The illumination, while dense, is composed of discrete, artificial LEDs, which cannot fully replicate the continuous, full spectrum, and High Dynamic Range (HDR) nature of global illumination in an outdoor environment (i.e., the sun, sky, and indirect bounces from the entire surroundings). WildRelight, in contrast, sacrifices per light decomposition to capture the full complexity of natural, ``in-the-wild" illumination for scenes level relighting.

\paragraph{Neural Inverse Rendering and Relighting.}
The advent of NeRF~\cite{mildenhall2021nerf} has revolutionized 3D reconstruction and sparked a new wave of neural inverse rendering methods. These approaches \cite{zhao2024illuminerf,zhang2021nerfactor,boss2021nerd,zhang2021physg} extend the NeRF framework to decompose a scene into its intrinsic properties from multiple views. They successfully disentangle geometry, materials, and illumination, allowing for high quality novel view synthesis and relighting. However, these methods either require controlled laboratory capture with known lighting~\cite{zhang2021physg} or attempt to estimate a single, static illumination (e.g., an environment map) from the multi-view images themselves~\cite{zhang2021nerfactor}. While impressive, they do not address the challenge of relighting under diverse, measured real-world illumination conditions.

\paragraph{Multi-View In-the-Wild Datasets.}
To move neural inverse rendering outdoors, recent datasets have successfully captured scenes under time-varying natural light. The NeRF-OSR~\cite{rudnev2022nerf} dataset captured time lapse videos of buildings to train a NeRF model that can relight the scene by interpolating the learned illuminations. Similarly, the Objects With Lighting (OWL)~\cite{Ummenhofer2024OWL} and Stanford-ORB~\cite{kuang2023stanford} datasets capture objects from multiple viewpoints under several distinct natural lighting conditions, providing GT envmaps for each. The fundamental design goal of these datasets is to provide \textit{multi-view} data for 3D reconstruction to build a relightable 3D model of the scene. Consequently, their data structure and benchmarks are built around training and testing multi-view reconstruction algorithms (e.g., NeRF, 3DGS) ~\cite{zhang2021nerfactor,toschi2023relight,kerbl3Dgaussians}. They are unsuitable for the distinct and equally challenging task of \textit{single-image} relighting, which assumes only one input photograph and no multi-view correspondence.

\paragraph{Classical and Single-View Datasets.}
The final category of related work consists of datasets that, like ours, are captured from a fixed, single viewpoint. However, these datasets were designed for different tasks and have critical limitations. Classical photometric stereo datasets~\cite{shi2016benchmark} capture objects under a sparse set of discrete point lights, typically in a darkroom, which is far removed from real world illumination. Other datasets designed for image processing tasks, such as the flash/no-flash dataset~\cite{aksoy2018dataset,murmann2019dataset}, provide only two simple, low dynamic range illumination conditions. While these datasets are valuable for their intended purpose, none provide the necessary data to evaluate modern, physically-based relighting algorithms: a collection of high resolution, HDR images of complex scenes, with each image rigorously paired with a spatially aligned, HDR GT envmap. WildRelight is the first dataset to fill this crucial void, bridging the gap between single-image generative models and the physical reality of our world.

\begin{figure*}[!t]
    \centering
    \includegraphics[width=0.8\linewidth]{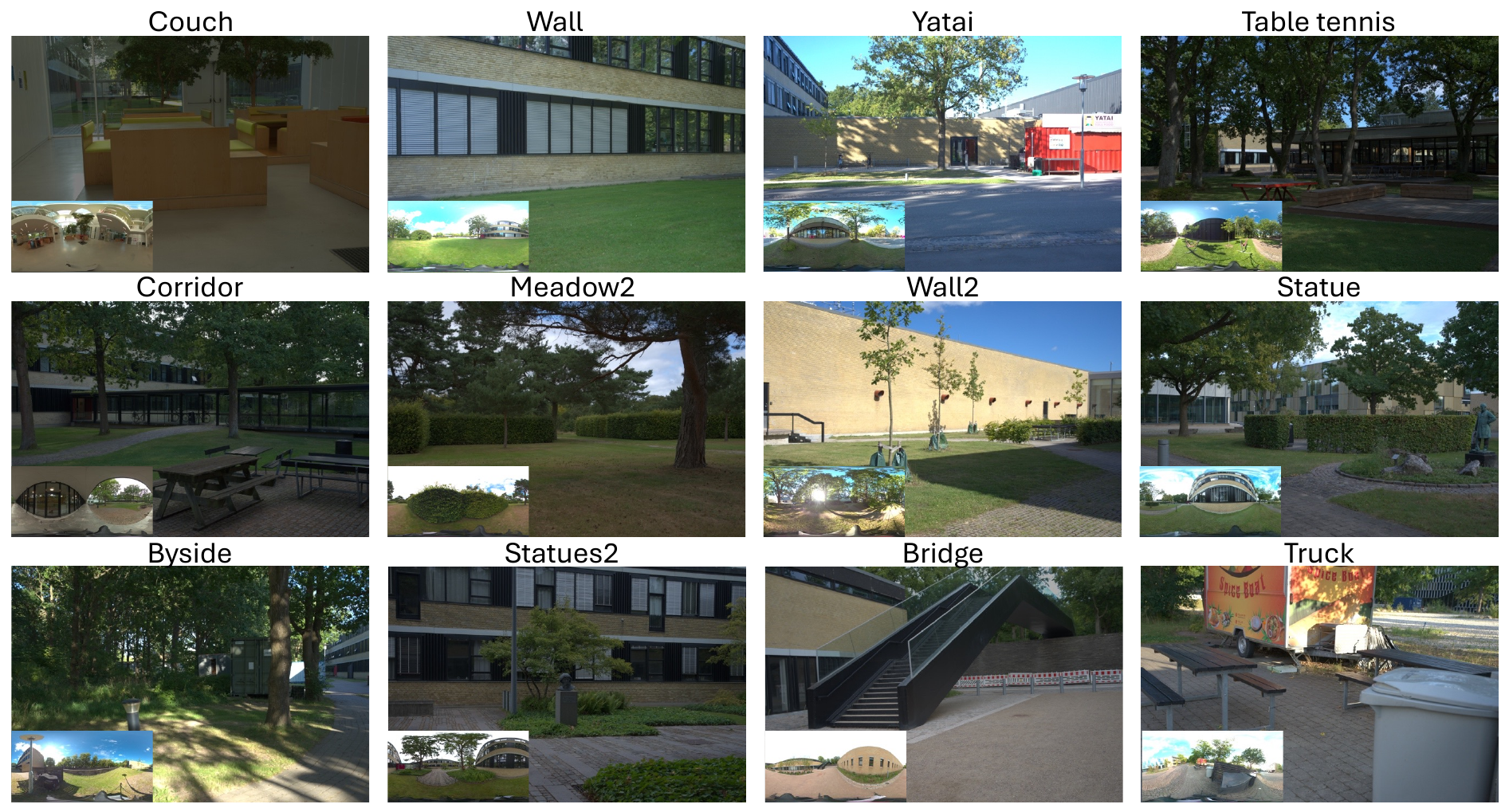}
    \caption{We selected some example scenes from the dataset. We meticulously curated the dataset to represent a diverse range of challenging scenarios. The collected scenes encompass complex environmental conditions including tree structures, transparent glass surfaces, and reflective glass materials. For each scene, temporal variations are captured through photographs taken at different time intervals, accompanied by corresponding environmental maps that provide spatial lighting information.}
    \label{fig:showcase}
    \vspace{-5ex}
\end{figure*}

\section{Dataset Collection and Curation}
\label{sec:dataset_collection}

Following the principles of rigorous and reproducible data acquisition, we introduce the WildRelight dataset, specifically designed for single-image relighting under real-world, dynamic illumination. Our collection protocol is meticulously crafted to ensure high fidelity in both scene capture and environmental lighting measurement.

\subsection{Dataset Overview}
\label{subsec:WildRelight_overview}

The WildRelight dataset contains 30 distinct scenes. To capture the full spectrum of natural lighting changes, each scene was recorded from a fixed camera position under 5 to 7 different illumination conditions. This approach provides a challenging and realistic benchmark for evaluating single-image relighting algorithms. The core of our data collection strategy is to sample lighting at different times of day, capturing both the subtle shifts of afternoon light and the rapid, dramatic changes during sunset. %
This temporal variation provides a unique benchmark for evaluating robustness to continuous illumination changes.

It is worth noting that unlike active lighting setups used in indoor datasets which can capture thousands of frames per second, our acquisition is inherently constrained by the immutable trajectory of the sun. Capturing a single scene's full dynamic range requires hours of continuous monitoring rather than seconds.
However, this constraint is necessary to ensure the distinct authenticity of ``in-the-wild'' natural light, which cannot be simulated by momentary active illumination.

\subsection{Data Acquisition Protocol}
\label{subsec:acquisition_protocol}

Our data acquisition relies on a dual-camera system: a Sony A7 is used to capture the high resolution scene, while an Insta360 Pro 2 simultaneously records the full 360-degree environmental illumination map (envmap).

\textbf{Spatio-Temporal Alignment.} A critical aspect for ensuring the precise correspondence between a captured image and its lighting environment is the spatial alignment of the two cameras. We guarantee this by co-locating the optical center of the Insta360 Pro 2 with the nodal point (entrance pupil) of the Sony A7 lens. This setup ensures that the captured envmap accurately represents the complete incident light field at the exact vantage point of the scene camera. To minimize temporal discrepancies caused by changing natural light, we streamlined our capture process. For each data point, we first captured the envmap with the Insta360 Pro 2, followed immediately by the scene capture with the Sony A7. This swap was typically accomplished in under one minute, and in rare cases, up to two minutes to avoid transient scene elements like pedestrians.

\textbf{Nodal Point Alignment.} Achieving this precise co-location required us to first empirically determine the ``no-parallax point" (entrance pupil) of the specific Sony A7 lens and focal length used. We employed a standard panoramic photography methodology: camera was mounted on a specialized panoramic head, and its forward-backward position was iteratively adjusted. The correct position was identified when rotating (panning) the camera caused zero observable parallax shift between two aligned, depth separated vertical objects (e.g., a near lamppost and a distant utility pole). Once this pivot point was locked, the Insta360 Pro 2 was mounted on a vertical rig, aligning its optical center with this exact vertical axis. A detailed guide of this calibration procedure is provided in the supplementary material.

\begin{figure}[!tbh]
    \centering
    \vspace{-3ex}
    \begin{subfigure}[b]{0.35\textwidth}
        \includegraphics[width=\linewidth]{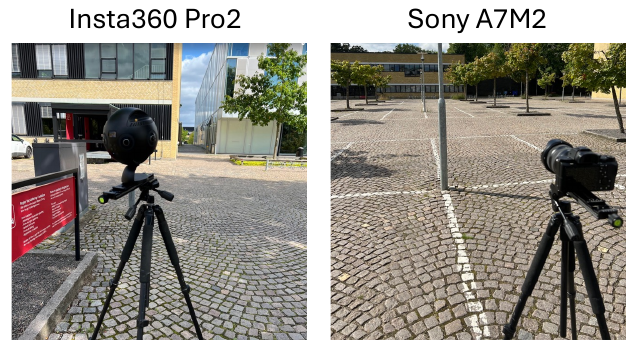}
        \caption{}
        \label{fig:capture_setting}
    \end{subfigure}
    \hfill
    \begin{subfigure}[b]{0.6\textwidth}
        \includegraphics[width=\linewidth]{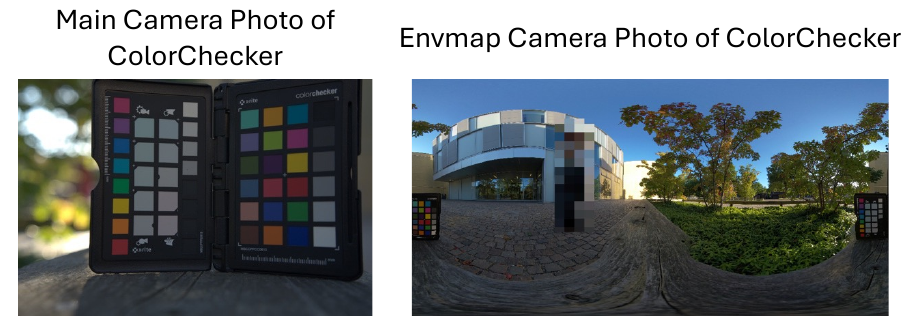}
        \caption{}
        \label{fig:xrite}
    \end{subfigure}
    \caption{(a) Capture settings. Both the Insta360 Pro2 and Sony A7M2 are mounted on a rail system to enable front-to-back adjustments for precise alignment with the nodal point. (b) Photos of Xrite ColorChecker are used to calibrate color between main camera Sony A7 and envmap camera Insta360 Pro2}
    \label{fig:combined}
    \vspace{-3ex}
\end{figure}

\begin{wrapfigure}{r}{0.45\textwidth}
    \centering
    \includegraphics[width=\linewidth]{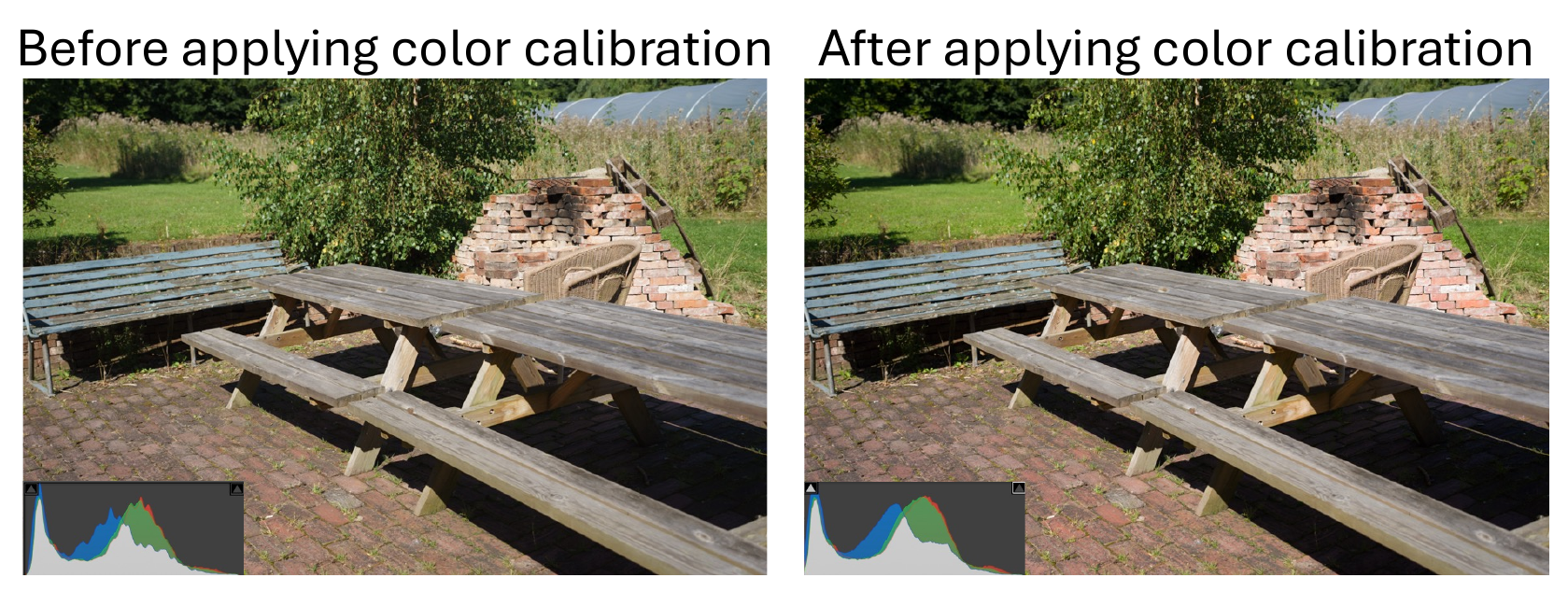}
    \caption{Comparison examples of color calibration effects. Due to the inherently sophisticated color science in professional grade products such as the Sony A7 and Insta360 Pro2, visual differences between pre- and post-color calibration images are negligible. To quantitatively analyze the calibration effects, a histogram was incorporated into the lower left corner of the image. While visual inspection reveals minimal variation, the histogram demonstrates discernible differences in pixel distribution before and after color calibration.}
    \label{fig:color_calibre}
    
\end{wrapfigure}

\textbf{Temporal Sampling Strategy.} Recognizing that natural light evolves at a non-uniform rate, we adopted a variable temporal sampling strategy. During the afternoon (11:00 - 17:00), when illumination changes are gradual, data was collected every 45 to 60 minutes. In contrast, during the pre-sunset period, when light intensity and chromaticity shift dramatically, we increased the capture frequency to every 10 to 15 minutes.

\textbf{Camera Parameters.} To maintain consistency, we used fixed camera settings where possible. For the Sony A7, we set the ISO to 100, aperture to f/4, and white balance to 5000K. A 40mm focal length was chosen for its natural field of view and minimal distortion, making it ideal for a general purpose dataset. The shutter speed was typically set to 1/500s in the afternoon and 1/100s in the evening. For scenes with extremely high dynamic range, we captured bracketed exposures by adjusting the shutter speed. The Insta360 Pro 2 was configured with matching ISO (100) and white balance (5000K), and its shutter speed was synchronized with the Sony A7. All images were captured in the 16 bit linear RAW format.

\subsection{Color and Radiometric Calibration}
\label{subsec:color_calibration}

To create a radiometrically accurate and color consistent dataset, we implemented a careful calibration and processing pipeline.

\textbf{Cross-Camera Color Calibration.} To correct for the inherent color discrepancies between the Sony and Insta360 sensors, we performed a one-time color calibration. Under diffuse, overcast daylight, we photographed an X-Rite ColorChecker target simultaneously with both cameras. Using the ColorChecker's accompanying software, we generated custom color profiles for each camera. These profiles were then applied during post-processing to all images in the dataset, unifying the color rendition across both capture devices.

\begin{wrapfigure}{r}{0.45\textwidth}
    \centering
    \includegraphics[width=0.9\linewidth]{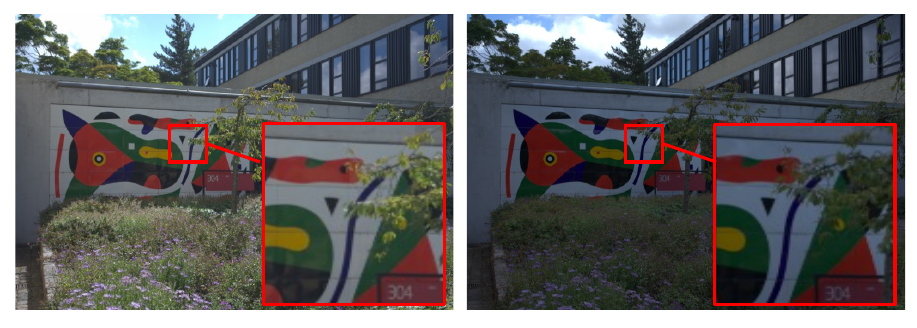}
    \caption{Example of Dynamic Scene Elements.
    As showed in the figure, the left and right images were captured at different times with a fixed camera position. Despite the static camera setup, subtle movements of the leaves occur due to external factors such as wind. To address this, we manually created masks for these dynamic regions, allowing researchers to determine whether to include them when computing metrics.}
    \label{fig:example_dynamic}
\end{wrapfigure}

\textbf{Linear HDR Synthesis.}
Our entire pipeline operates in a linear color space, starting from the RAW sensor data. By using RAW files, we circumvent the need for estimating a non-linear camera response function (CRF), a common source of error when working with processed image formats like JPEG.
For bracketed captures, we merge the multiple exposures into a single, HDR image. Given the linearity of the RAW data, the radiance $L_i$ from a single exposure $i$ is directly proportional to its recorded linear pixel value $Z_i$ (scaled to $[0, 1]$) divided by the exposure time $\Delta t_i$.

We merge bracketed exposures by computing a weighted average of radiance $L_i = Z_i / \Delta t_i$, where $Z_i$ is the normalized pixel value and $\Delta t_i$ is the exposure time. To mitigate noise and saturation, we employ a triangular weighting function~\cite{10.1145/258734.258884} $w(z) = 1.0 - |2z - 1.0|$:$$L = \frac{\sum_{i} w(Z_i) \cdot (Z_i / \Delta t_i)}{\sum_{i} w(Z_i)}$$The final HDR radiance is accumulated in high-precision float64 and stored in linear EXR format to ensure faithful physical representation.

\subsection{Dynamic Element Masking}
A significant challenge in our longitudinal, ``in-the-wild" capture is the inevitable presence of dynamic scene elements, such as wind-blown foliage and moving cloud formations, despite a static camera rig. To preserve the photometric integrity of our GT images, we avoid computational alignment (e.g., warping) which would alter the pixel data. Instead, we provide meticulously hand-annotated binary masks for all non-static regions. This approach allows researchers to optionally exclude these dynamic areas during metric computation, thereby isolating the evaluation of relighting performance from artifacts caused by scene motion. After determining that automated methods were unreliable for our complex natural scenes, we developed a rigorous manual annotation pipeline based on pairwise temporal comparisons. The complete details of this pipeline, including our annotation interface and specific exclusion criteria (e.g., for water and reflections), are available in the supplementary materials.

\section{Methodology}
\label{sec:proposed_methodology}

To illustrate how WildRelight’s real-world supervision can be practically exploited, we design a reference framework that integrates physics-guided posterior sampling for inverse decomposition regularization, together with sampling-aware TTA to better align forward relighting dynamics.

Rather than aiming to introduce a fully optimized solution, this framework serves as a structured case study demonstrating how dataset-driven supervision can be incorporated at both inference and adaptation stages. The resulting bidirectional consistency between scene representation and image formation highlights the practical value of WildRelight, while leaving substantial room for future methodological improvements.

\subsection{Physics-Guided Inverse Rendering via Diffusion Posterior Sampling}
\label{sec:method_dps}
To enforce physical validity in G-buffer prediction without retraining, we introduce a physics-guided inference strategy based on Diffusion Posterior Sampling  ~\cite{chung2023diffusion}.
At diffusion timestep $t$, the estimated clean latent $\hat{x}_0$ is decoded by the VAE $D(\cdot)$ into intrinsic components and rendered under illumination $L$ via a differentiable Cook--Torrance renderer $\mathcal{R}(\cdot)$~\cite{cook1982reflectance,walter2007microfacet}.
We enforce consistency with the observed image $I_{\text{gt}}$ via a measurement loss:
\begin{equation}
\mathcal{L}_{\text{render}} = \left\| \mathcal{R}\big(D(\hat{x}_0),\, L\big) - I_{\text{gt}} \right\|_2^2 .
\end{equation}
Guided by the gradient $g_t = \nabla_{x_t} \mathcal{L}_{\text{render}}$, the DDIM sampling trajectory is refined as:
\begin{equation}
x_{t-1} \leftarrow x_{t-1} - \zeta_t\, g_t ,
\end{equation}
where $\zeta_t$ is the guidance strength.
To ensure stability and efficiency during this optimization, we employ a split-sum approximation for image-based lighting while keeping the diffusion network parameters frozen.

\subsection{Forward Relighting via Sampling-Aware Temporal Test-Time Adaptation}
\label{sec:method_tta}

Standard models struggle with real-world domain shifts. We propose a TTA framework that leverages the temporal illumination variations in WildRelight for self-supervision.

\paragraph{Temporal Self-Supervision.} Adopting a leave-one-out protocol, we use $N-1$ observed lighting conditions to adapt the model to the specific scene, testing on the held-out light. This relies strictly on paired (photo, envmap) data without GT G-buffers.

\paragraph{Parameter-Efficient Adaptation.} We freeze the diffusion backbone and optimize only lightweight LoRA modules~\cite{hu2022lora} injected into the attention layers, preventing overfitting on the limited per-scene data.

\paragraph{Sampling-Aware Optimization.} Instead of standard noise prediction, we optimize the \emph{denoising trajectory}. We execute a partial DDIM sampling and backpropagate a perceptual reconstruction loss through the final $K$ steps: $\mathcal{L} = \mathcal{L}_{\text{noise}} + \lambda_{\text{perc}} \mathcal{L}_{\text{LPIPS}}(D(z_0), I_{\text{target}})$. This implicitly aligns the renderer's dynamics with the scene's specific light transport characteristics (e.g., complex shadows) efficiently.

\section{Experiments}
\label{sec:experiments}

We validate \textit{WildRelight} and our proposed methods in three phases: benchmarking zero-shot SOTA methods to quantify the synthetic-to-real gap, supervised finetuning to demonstrate dataset utility, and evaluating our proposed DPS and TTA strategies (Sec.~\ref{sec:eval_methodology}).

\paragraph{Evaluation Protocol.}
All experiments use standard metrics (PSNR, SSIM, LPIPS) computed under a single, shared alignment procedure, but they differ in which portion of \textit{WildRelight} they are evaluated on, and we make this explicit because it accounts for the different zero-shot baseline values reported across \autoref{tab:baseline_setups}, \autoref{tab:baseline_finetune} and \autoref{tab:ablation_method}.
The zero-shot benchmark (\autoref{tab:baseline_setups}) is aggregated over all 30 scenes; the supervised finetuning experiment (\autoref{tab:baseline_finetune}) necessarily reports on the 5-scene held-out test split only; and our inference-time methodology (\autoref{tab:ablation_method}), which requires no training split, is again evaluated over all 30 scenes.
The two full-dataset protocols also differ in how (input, target) pairs are formed: the zero-shot benchmark decomposes a single fixed ``time 0'' photograph per scene and relights it to the remaining $N-1$ illuminations, whereas the leave-one-lighting-out protocol of Sec.~\ref{sec:eval_methodology} lets every one of the $N$ illuminations serve as the held-out target in turn. The resulting pair sets are therefore not identical, which is why the same pre-trained DiffusionRenderer scores 22.81 dB in \autoref{tab:baseline_setups} and 21.63 dB in \autoref{tab:ablation_method}.
A critical challenge in single-image relighting is the inherent scale ambiguity. To ensure rigorous evaluation, we adopt a global least-squares alignment strategy for all experiments. Specifically, we solve for an optimal scalar $\alpha$ to align the global intensity of predictions with the ground truth before metric computation. Mathematical derivations and implementation details are provided in the supplementary material.

\subsection{Benchmarking SOTA: The Sim-to-Real Gap}
\label{sec:benchmark_sota}

To quantify the domain gap, we evaluate RGB$\leftrightarrow$X~\cite{zeng2024rgb}, DiffusionRenderer~\cite{liang2025diffusion}, and Materialist~\cite{wang2026materialist} (see \autoref{tab:baseline_setups}, Zero-Shot row).
Both generative models trained on synthetic data fail to capture complex high-frequency shadows and outdoor indirect illumination, but to very different degrees. The failure is severe for RGB$\leftrightarrow$X ($15.87$ dB PSNR, $0.4507$ SSIM), which does not recover the global intensity of outdoor scenes at all. DiffusionRenderer is markedly stronger ($22.81$ dB, $0.6218$ SSIM), yet its outputs remain perceptually far from the ground truth ($0.3927$ LPIPS), and \autoref{tab:baseline_finetune} shows that the same architecture gains a further $2.67$ dB on the held-out test split once it is finetuned on \textit{WildRelight}. This explicitly highlights the Sim-to-Real Gap: synthetic priors do not generalize zero-shot to in-the-wild environments.
Optimization-based Materialist achieves relatively higher metrics because its protocol utilizes the \textit{known} GT environment map during optimization (see Suppl. for protocol details). However, its reliance on explicit geometry optimization makes it less flexible than generative approaches, often exhibiting artifacts in complex vegetation.

\begin{figure*}
    \centering
    \includegraphics[width=\linewidth]{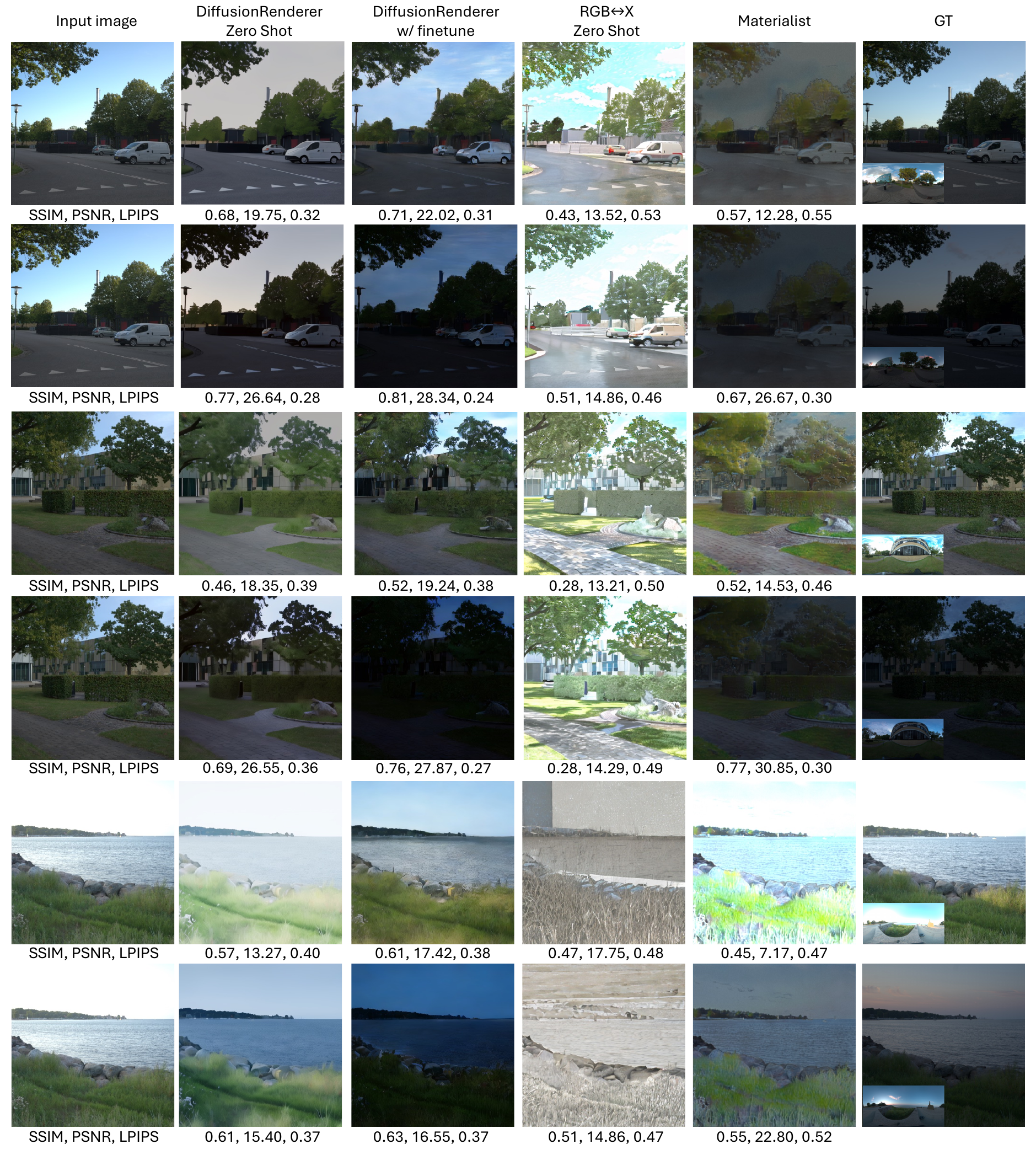}
    \caption{Qualitative Comparison of Different Methods on the Test Set Relighting. The numerical values displayed beneath the image correspond to the following metrics: SSIM, PSNR, and LPIPS. In zero shot scenarios, both DiffusionRenderer \cite{liang2025diffusion} and RGB$\leftrightarrow$X \cite{zeng2024rgb} struggle to accurately render image brightness. After finetuning, DiffusionRenderer demonstrates improved alignment with GT in the test set, achieving more accurate brightness reproduction. Materialist \cite{wang2026materialist}, leveraging an optimization-based approach combined with a physical renderer, though exhibits good brightness representation in rendered images, its reliance on geometric accuracy and the inherent limitations of physical renderers lead to unrealistic outputs compared to Diffusion Model-based methods. The envmap used for relighting is in the GT column (lower left).}
    \label{fig:baseline_compare1}
\end{figure*}

\subsection{Dataset Utility: Supervised Finetuning}
\label{sec:dataset_utility}

\begin{table}[!tb]
    \centering
    \begin{minipage}[t]{0.48\textwidth}
        \centering
        \captionof{table}{Performance of pretrained baseline models, aggregated over all 30 scenes of \textit{WildRelight}. Each scene is decomposed from its fixed ``time 0'' photograph and relit to the remaining $N-1$ illuminations.}
        \label{tab:baseline_setups}
        \resizebox{\linewidth}{!}{%
            \setlength{\tabcolsep}{1mm}
            \begin{tabular}{l c c c}
            \toprule
            Method & PSNR ($\uparrow$) & SSIM ($\uparrow$) & LPIPS ($\downarrow$) \\
            \midrule
            \multicolumn{4}{l}{\textit{Zero-Shot Neural Relight}} \\
            \quad RGB$\leftrightarrow$X~\cite{zeng2024rgb} & 15.87 & 0.4507 & 0.4917 \\
            \quad DiffRender~\cite{liang2025diffusion} & 22.81 & 0.6218 & 0.3927 \\
            \midrule
            \multicolumn{4}{l}{\textit{Opt w/ GT Illumination}} \\
            \quad Materialist~\cite{wang2026materialist} & 24.19 & 0.5819 & 0.3639 \\
            \bottomrule
            \end{tabular}
        }
    \end{minipage}
    \hfill
    \begin{minipage}[t]{0.48\textwidth}
        \centering
        \captionof{table}{Effect of finetuning DiffusionRenderer \cite{liang2025diffusion} on \textit{WildRelight}. Both rows are evaluated on the 5-scene held-out test split, so the zero-shot row differs from the all-30-scene value in \autoref{tab:baseline_setups}.}
        \label{tab:baseline_finetune}
        \resizebox{\linewidth}{!}{%
            \begin{tabular}{l c c c}
            \toprule
            Model & PSNR ($\uparrow$) & SSIM ($\uparrow$) &  LPIPS ($\downarrow$) \\
            \midrule
            Zero-shot & 23.28 & 0.6165 & 0.3790 \\
            Finetuned & 25.95 & 0.6687 &  0.3368\\
            \bottomrule
            \end{tabular}
        }
    \end{minipage}
    \vspace{-4ex}
\end{table}

To validate the efficacy of \textit{WildRelight} in bridging the sim-to-real gap, and to establish a supervised performance upper bound, we conducted a global finetuning experiment.
We adopted the state-of-the-art DiffusionRenderer~\cite{liang2025diffusion} as our base model.
Given the prohibitive cost of full-parameter retraining, we employed LoRA~\cite{hu2022lora} to efficiently adapt the pre-trained UNet to the lighting and material statistics of our 22-scene training set, using the 3-scene validation set for checkpoint selection and reporting on the 5-scene held-out test split.
The VAE and environment encoder remained frozen to preserve their generative priors.
Detailed training configurations and architectural specifications are provided in the supplementary material.

\paragraph{Results and Analysis.}
As reported in \autoref{tab:baseline_finetune}, supervised adaptation on \textit{WildRelight} yields a dramatic performance boost over the zero-shot baseline.
The PSNR improves from 23.28 dB to 25.95 dB, with consistent gains in SSIM and LPIPS.
This substantial improvement confirms two critical findings:
(1) There exists a significant domain gap between current synthetic training data and real-world outdoor scenes;
(2) \textit{WildRelight} contains the necessary high-fidelity signals to effectively bridge this gap, serving as a vital resource for future data-driven approaches.
This finetuned model serves as a strong \textit{supervised reference} for our subsequent evaluation of zero-shot methodologies.

\begin{figure*}[!t]
    \centering
    \includegraphics[width=0.9\linewidth]{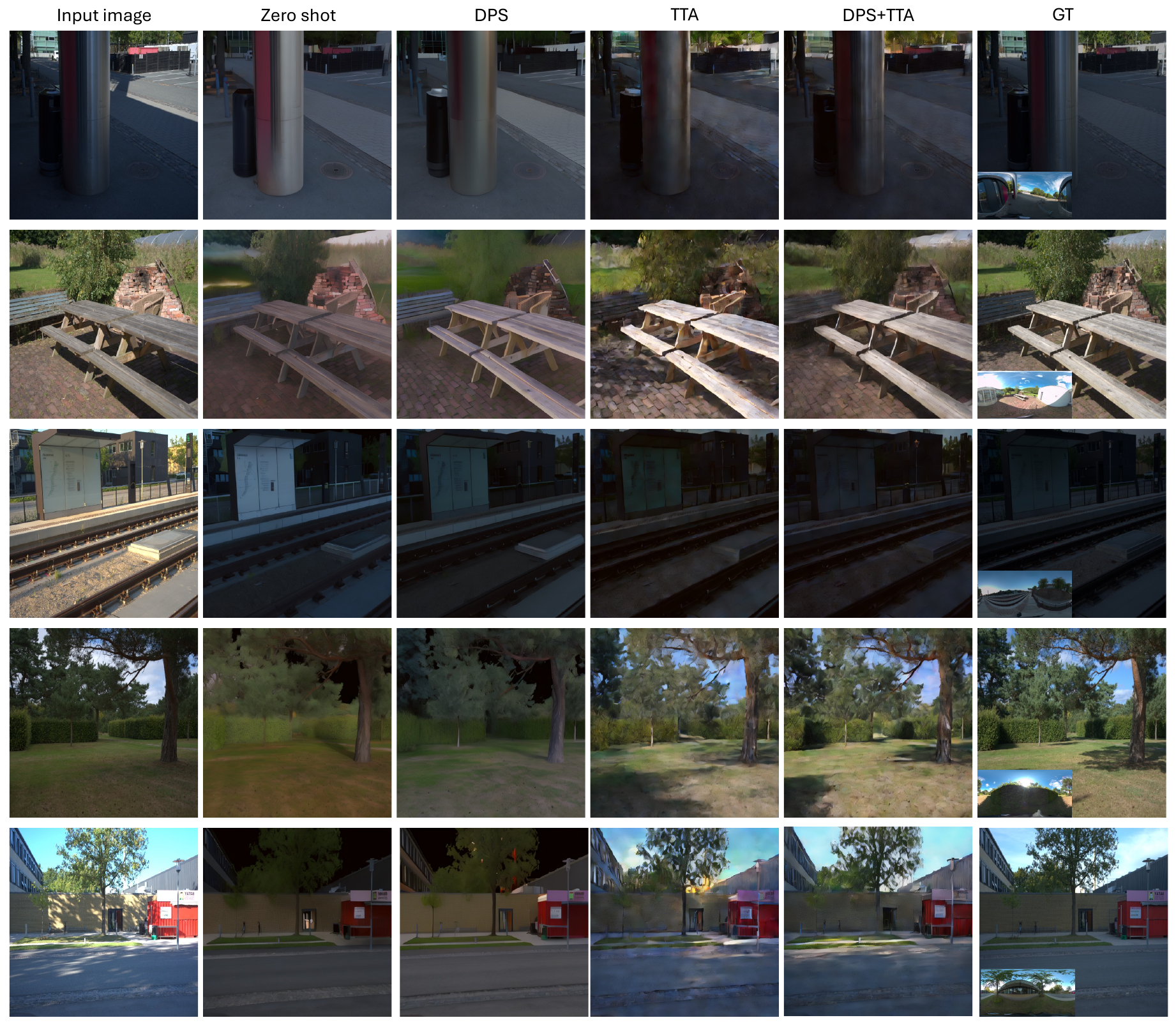}
    \caption{Qualitative Ablation Study Results.
        We visualize the results of our proposed framework on the \textit{WildRelight} dataset. The envmap on bottom left of GT image correspond to target illuminations used for relighting.
        }
    \label{fig:ours_results}
\end{figure*}

\subsection{Evaluation of Proposed Methodology}
\label{sec:eval_methodology}

\paragraph{Evaluation Protocol.}
Unlike global finetuning which relies on fixed splits, our inference-time methodology (DPS + TTA) enables instance-specific adaptation. We evaluate this universality across all 30 scenes using a ``leave-one-lighting-out'' protocol: for $N$ lightings, one is the test target while the other $N-1$ serve as self-supervised signals. This rigorously simulates real-world deployment without ground-truth supervision.

\begin{table}[t]
    \centering
    \caption{
        \textbf{Ablation Study of the Proposed Methodology.} 
        Evaluated on the full 30-scene \textit{WildRelight} dataset under the leave-one-lighting-out protocol; the baseline row therefore differs from the fixed ``time 0'' input protocol of \autoref{tab:baseline_setups}.
        While TTA alone significantly improves PSNR, it inadvertently degrades perceptual quality (higher LPIPS) due to unconstrained optimization overfitting to pixel-level intensities. 
        Incorporating DPS restores physical consistency, and their combination yields the best overall performance, effectively bridging the sim-to-real gap without supervised training.
    }
    \label{tab:ablation_method}
        \begin{tabular}{l l c c c}
            \toprule
            Configuration & Mechanism & PSNR $\uparrow$ & SSIM $\uparrow$ & LPIPS $\downarrow$ \\
            \midrule
            Baseline (Pre-trained) & Zero-shot & 21.63 & 0.6311 & 0.3901 \\
            \quad + DPS & Inference Prior & 22.58 & 0.6578 & 0.3825 \\
            \quad + TTA & Optimization & 24.10 & 0.6451 & 0.3923 \\
            \midrule
            \textbf{\quad + DPS \& TTA} & \textbf{Constrained Adapt.} & \textbf{25.04} & \textbf{0.6829} & \textbf{0.3453} \\
            \bottomrule
        \end{tabular}%
    \vspace{-2ex}
\end{table}

\subsubsection{Ablation Study and Analysis}
\label{sec:ablation_results}

We dissect the contribution of each component by incrementally integrating them into the pre-trained DiffusionRenderer~\cite{liang2025diffusion} baseline. The quantitative and qualitative results, averaged over the full dataset, are presented in \autoref{tab:ablation_method} and \autoref{fig:ours_results}.

\paragraph{Baseline (Zero-Shot).}
The synthetic-trained baseline suffers a substantial sim-to-real gap (21.63 dB PSNR), failing to disentangle complex outdoor illumination and reflectance. This deficit quantifies the limitations of synthetic data rather than the architecture. We use this baseline to represent synthetic priors, demonstrating that real-world adaptation—enabled by our data—is critical for bridging this gap.

\paragraph{Impact of Physics-Guided DPS.} 
Introducing Diffusion Posterior Sampling (+DPS) injects physical constraints, yielding consistent gains (+0.95 dB PSNR). Crucially, DPS acts as a geometric anchor: by enforcing rendering equation consistency, it prevents hallucinated shadows, ensuring generation adheres to the underlying scene structure.

\paragraph{Impact of Temporal TTA.} 
TTA alone drastically minimizes photometric error (+2.5 dB PSNR) but slightly degrades LPIPS ($0.390 \rightarrow 0.392$). This reveals a photometric-perceptual trade-off in unconstrained self-supervision: optimization overfits pixel intensities at the cost of the natural image manifold, leading to artifacts that penalize perceptual metrics.

\paragraph{Synergy: Constrained Adaptation.} 
The full pipeline (+DPS \& TTA) resolves this trade-off, achieving the best overall performance (25.04 dB PSNR, 0.345 LPIPS). Here, DPS regularizes the TTA trajectory, ensuring the model adapts to scene-specific lighting dynamics without sacrificing physical plausibility or high-frequency details.

\paragraph{Comparison with Global Finetuning.}
Remarkably, our inference-time approach rivals fully supervised global finetuning reported in Sec.~\ref{sec:dataset_utility} (25.04 vs. 25.95 dB), though the two figures are measured under different protocols (all 30 scenes under leave-one-lighting-out vs. the 5-scene held-out test split; cf. the Evaluation Protocol of Sec.~\ref{sec:experiments}) and are therefore indicative rather than a like-for-like comparison. This underscores our framework's efficiency: effectively bridging the sim-to-real gap with near-supervised performance without expensive retraining.

\section{Conclusion}
In this work, we introduce \textbf{WildRelight}, a benchmark bridging the synthetic-to-real gap via aligned, in-the-wild images and co-located HDR envmaps. 
Crucially, the dataset's temporal nature enables reformulating domain adaptation as a real-world self-supervised task. Our proposed Physics-Guided Inverse Rendering (DPS) and Sampling-Aware TTA validate this, achieving performance rivaling fully supervised baselines without synthetic pre-training.
Future directions include modeling dynamic scene elements rather than masking them, and optimizing the computational efficiency of instance-specific adaptation. We hope WildRelight catalyzes the development of models physically robust to the unconstrained complexity of the natural world.
\clearpage
\bibliographystyle{splncs04}
\bibliography{references}

\clearpage
\setcounter{page}{1}
\setcounter{section}{0}

\section*{\huge Supplementary Materials}

\section{Quantitative Validation of Illumination Alignment}
\label{sec:alignment_validation}

We provide a rigorous quantitative validation based on metadata timestamp statistics and solar angular displacement analysis. This proves that the temporal gap in our acquisition pipeline results in physically negligible illumination misalignment for the task of relighting.

\subsection{Temporal Synchronization Statistics}

We extracted and analyzed the acquisition timestamps from the metadata of all image pairs (Sony A7 scene images and Insta360 envmaps). The distribution of the time delay $\Delta t$ is summarized in Table~\ref{tab:time_delta}.

\begin{table}[h]
    \centering
    \caption{Statistics of Temporal Synchronization ($\Delta t$) between scene and environment map capture.}
    \label{tab:time_delta}
    \begin{tabular}{lc}
        \toprule
        \textbf{Metric} & \textbf{Value} \\
        \midrule
        Median Time Delta & 38.00 seconds \\
        Mean Time Delta & 40.14 seconds \\
        Max Time Delta & 114.00 seconds \\
        \bottomrule
    \end{tabular}
\end{table}

While the maximum delay is approximately 1.7 minutes, we demonstrate below that this is well within the tolerance for accurate outdoor lighting estimation.

\subsection{Physical Error Analysis: Solar Angular Displacement}

The primary source of directional illumination outdoors is the sun. The Earth rotates at approximately $15^\circ$ per hour ($0.00417^\circ$ per second). We calculate the angular displacement of the sun, $\theta_{\text{shift}}$, caused by the capture delay:

\begin{equation}
    \theta_{\text{shift}} = \Delta t \times 0.00417^\circ/s
\end{equation}

For our mean delay (40.14s):
\begin{equation}
    \theta_{\text{mean}} \approx 0.17^\circ
\end{equation}

For our worst-case delay (114s):
\begin{equation}
    \theta_{\text{max}} \approx 0.48^\circ
\end{equation}

To contextualize these values, the angular diameter of the sun is approximately $0.5^\circ$. 
In the average case, the sun moves only $1/5$ of its own diameter, which is perceptually imperceptible in lighting effects.
Even in the worst-case scenario, the displacement ($0.48^\circ$) is still less than the angular size of the light source itself ($0.5^\circ$).

\begin{figure}[tb!]
    \centering
    \includegraphics[width=0.5\linewidth]{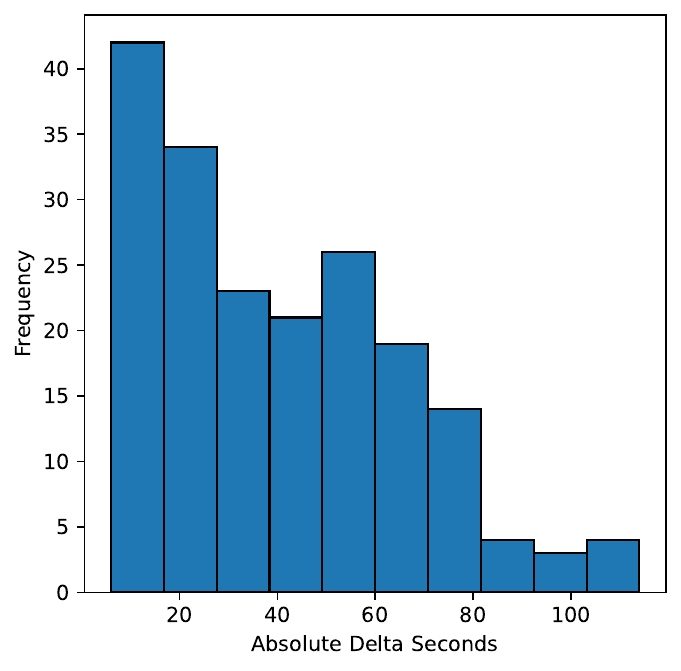}
    \caption{Distribution of capture time differences ($\Delta t$). The histogram shows the frequency of absolute time delays between the scene image and the environment map capture across the dataset. The distribution is heavily right-skewed, with the vast majority of samples having a delay of less than 20 seconds, confirming that large delays are rare.}
    \label{fig:time_delta_hist}
\end{figure}

\subsection{Impact on Relighting Tasks}

Modern single-image relighting algorithms typically operate on environment maps that are downsampled (e.g., to $256 \times 128$) to estimate lighting coefficients (like Spherical Harmonics) or to condition diffusion models.

At a width of 256 pixels, the horizontal angular resolution is $360^\circ / 256 \approx 1.4^\circ$ per pixel.
\begin{equation}
    \text{Pixel Shift}_{\text{max}} = \frac{0.48^\circ}{1.4^\circ/\text{pixel}} \approx 0.3 \text{ pixels}
\end{equation}

Thus, even our maximum temporal delay results in a sub-pixel shift (0.3 pixels) in the effective resolution used by state-of-the-art relighting models. This quantitatively confirms that our dataset maintains high-quality illumination alignment suitable for training and evaluating inverse rendering methods.

We also plot the distribution of the absolute time differences ($\Delta t$) in Figure~\ref{fig:time_delta_hist}. 
As illustrated, the distribution is strongly skewed towards zero. The dominant peak in the first bin ($<20$s) corroborates that for the majority of our collected scenes, the temporal gap is minimal. The long tail extending to $\sim100$s represents a small fraction of outlier cases. 
Combined with the solar angular displacement analysis, this distribution confirms that the impact of these delays on illumination alignment remains physically negligible for the purpose of relighting.

\section{Details Setting of Baseline Benchmark}
\label{sec:baseline_methods}

\paragraph{Dataset and Metrics}
For the supervised finetuning experiment, our \textit{WildRelight} dataset is partitioned into a 22-scene training set, a 3-scene validation set, and a 5-scene hold-out test set. This split is used only where training is involved: the zero-shot baseline benchmark (\autoref{tab:baseline_setups}) requires no training and is therefore aggregated over all 30 scenes, as is the leave-one-lighting-out evaluation of our inference-time methodology (\autoref{tab:ablation_method}). We evaluate performance using three standard image quality metrics: Peak Signal to Noise Ratio (PSNR), the Structural Similarity Index (SSIM), and the Learned Perceptual Image Patch Similarity (LPIPS).

\paragraph{Baseline Models}
We selected three representative methods that support single-image relighting: RGB$\leftrightarrow$X~\cite{zeng2024rgb}, DiffusionRenderer~\cite{liang2025diffusion}, and Materialist~\cite{wang2026materialist}.

\paragraph{Evaluation Protocol (Diffusion-Based)}
The RGB$\leftrightarrow$X and DiffusionRenderer models share a similar architecture. They first employ a model trained on synthetic data to decompose a single input image into G-buffer components (albedo, roughness, metallic, normal). Then, they use a diffusion-based model as a neural renderer to relight the scene using these G-buffers and a novel envmap.

For our evaluation, we use the `time 0' photograph from each scene as the input to their publicly available, pretrained decomposition models. We then use the resulting G-buffers, along with the GT envmaps from the other time lapses in that scene, as input to their respective rerendering diffusion models. The rendered images are then compared against the corresponding GT photographs. Since the original GT photographs are provided in HDR linear color space, we first process them with tone mapping~\cite{reinhard2002photographic} followed by gamma correction ($\gamma \approx 2.2$) before computing the comparison metrics against the model outputs.

\paragraph{Evaluation Protocol (Optimization-Based)}
Materialist \cite{wang2026materialist} utilizes a different, optimization-based approach. It first predicts an initial G-buffer set, similar to the other methods. However, it then refines these G-buffers by optimizing them within a physical renderer (Mitsuba \cite{jakob2022mitsuba3}) to match the input image.
Given that our dataset provides high fidelity GT envmaps, we adapt their protocol to isolate material estimation. Instead of jointly optimizing for a coarse envmap, we provide the known GT envmap directly to the Mitsuba renderer during optimization. This allows the model to focus solely on optimizing the G-buffers (albedo, roughness, metallic, normal) to be physically consistent with the known input image and its corresponding lighting. For each scene, we provide two GT envmaps to facilitate this optimization. The resulting, refined G-buffers are then used to relight the scene with the remaining, previously unseen GT envmaps from that scene.

\paragraph{Finetuning Protocol}
To demonstrate the utility of \textit{Wild\-Relight} for domain adaptation, we finetuned the DiffusionRenderer~\cite{liang2025diffusion} model. Since the official training code was not publicly available, we implemented a custom finetuning pipeline based on its described architecture, which consists of a VAE, a dedicated environment encoder, and the core spatio temporal UNet.

For efficient adaptation, we froze the pre-trained weights of both the VAE and the environment encoder, isolating the finetuning process to the UNet. We employed Low-Rank Adaptation (LoRA)~\cite{hu2022lora} to update the UNet's weights. Specifically, we applied LoRA adapters with rank $r=8$ to the attention layers (to k,  to q, to v, to out.0) and the main projection layers (proj in, proj out).
During training, model operates in the VAE's latent space. The G-buffer components (basecolor, normal, depth, roughness, metallic) are individually encoded by the frozen VAE, and their resulting latents are concatenated to form the spatial conditioning input for the UNet. Concurrently, the GT HDR envmap is processed into three representations (LDR, log space, and a panoramic normal map). These are encoded by VAE, concatenated, passed through the frozen environment encoder to produce multi scale features, which are injected into UNet via cross attention.

The model was trained to predict the noise added to the target image's latent representation, following the DDPMScheduler sampling, and optimized using an $L_2$ (MSE) loss. We used the AdamW optimizer~\cite{loshchilov2017decoupled} with a learning rate of $1 \times 10^{-4}$ and a linear warmup schedule. The model was trained on our 22-scene training set for 48 hours on a single NVIDIA H100 GPU. The final checkpoint was selected based on the best performance on the 3-scene validation set, and all final metrics are reported on the 5-scene test set.

\paragraph{Results}
\label{sec:baseline_results}

The quantitative results of our baseline experiments are presented in \autoref{tab:baseline_setups} and \autoref{tab:baseline_finetune}.

\autoref{tab:baseline_setups} compares the zero shot performance of the three pre-trained models, aggregated over all 30 scenes of our dataset. As anticipated, the diffusion-based models (RGB$\leftrightarrow$X \cite{zeng2024rgb} and DiffusionRenderer \cite{liang2025diffusion}), which were trained exclusively on synthetic data, perform poorly on our real world images. This highlights a significant domain gap. The degradation is most severe for RGB$\leftrightarrow$X (15.87 dB PSNR, 0.4507 SSIM), which fails to recover the global intensity of outdoor scenes; DiffusionRenderer is markedly stronger (22.81 dB PSNR, 0.6218 SSIM) but remains perceptually far from the ground truth (0.3927 LPIPS), and \autoref{tab:baseline_finetune} shows it still has substantial headroom once finetuned on our training split. Materialist \cite{wang2026materialist} achieves better average metrics results. This is expected, as our protocol for this method provides the GT envmap during its optimization step, allowing it to produce more accurate, physically-based G-buffers. Additionally, the method's reliance on a physical renderer enables exceptional performance in low-light conditions, which contributes to the elevated average score. This result validates the efficacy of its optimization pipeline when illumination is known, but it is not a direct comparison to the zero shot relighting task faced by the other models.

\autoref{tab:baseline_finetune} illustrates the practical value of our dataset. By finetuning DiffusionRenderer on our training split, its performance on the unseen 5-scene test set improves dramatically. The PSNR increases from 23.28 dB to 25.95 dB, the SSIM score improves from 0.6165 to 0.6687, and LPIPS drops from 0.3790 to 0.3368. Note that this 23.28 dB zero-shot figure is measured on the held-out test split alone, and therefore differs from the 22.81 dB averaged over all 30 scenes in \autoref{tab:baseline_setups}. This significant boost demonstrates that \textit{WildRelight} provides a valuable resource for adapting existing models to the complex and diverse appearances of real-world materials and illumination, effectively bridging the synthetic-to-real domain gap.

\section{Evaluation Protocol and Scale Alignment.}
All experiments use the standard metrics PSNR, SSIM, and LPIPS, computed on the data portion specified for each experiment in Sec.~\ref{sec:baseline_methods}: the 5-scene test split for supervised finetuning, and all 30 scenes for the zero-shot benchmark and for our inference-time methodology.
A critical challenge in single-image relighting is the inherent \textit{scale ambiguity}—the absolute magnitude of illumination cannot be uniquely determined from a single observation. Consequently, direct comparisons between predicted and ground-truth (GT) intensities are often biased by arbitrary global scaling. 
To ensure a fair and rigorous evaluation across \textbf{all} reported experiments (including baselines, finetuning, and our method), we adopt a global least-squares alignment strategy. For every predicted image $\mathbf{I}_{pred}$ and its corresponding ground truth $\mathbf{I}_{gt}$, we solve for an optimal scalar $\alpha$:
\begin{equation}
    \alpha^* = \operatorname*{argmin}_\alpha \left\| \mathbf{I}_{pred} \cdot \alpha - \mathbf{I}_{gt} \right\|^2.
\end{equation}
Metrics are computed on the aligned prediction $\mathbf{I}_{pred} \cdot \alpha^*$. This protocol compensates for global intensity discrepancies while strictly preserving the relative illumination structure and chromaticity, thereby enabling a reliable assessment of relighting quality.

\section{Advantages of RAW-Based HDR Image}
\label{sec:raw}
\begin{figure*}[tbh!]
    \centering
    \includegraphics[width=\linewidth]{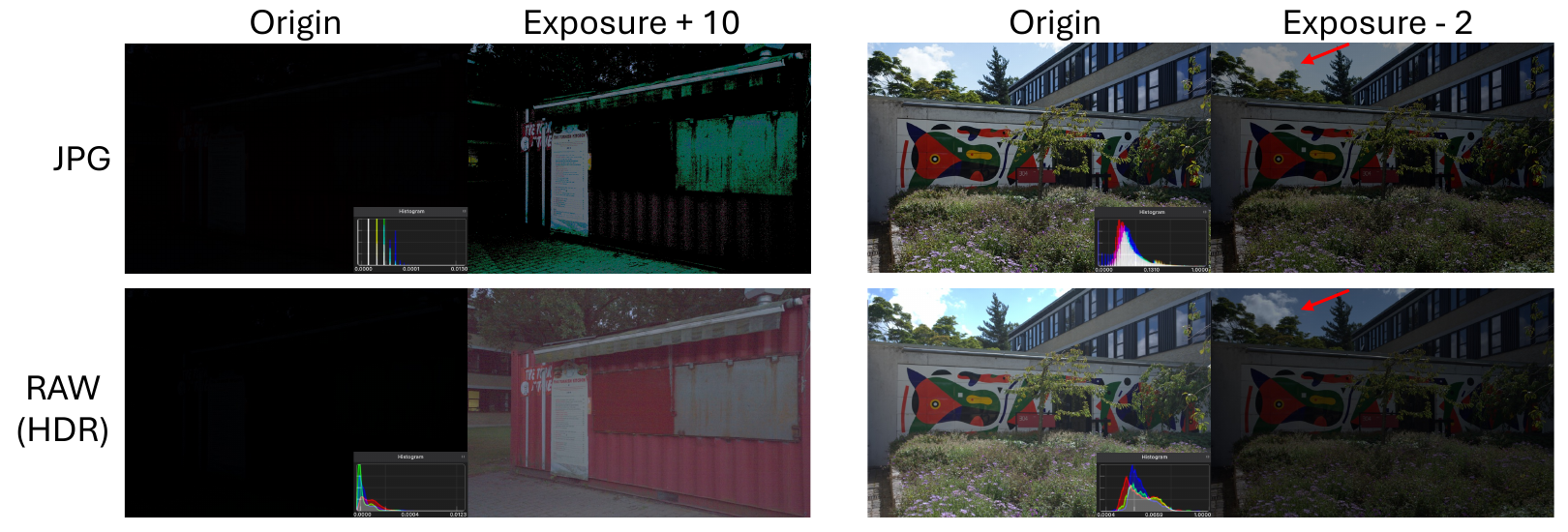}
    \caption{Advantage of RAW format with HDR. When using HDR photo, the details of the photo are preserved. Therefore, by adjusting the exposure settings, the original colors of the photo can be accurately restored. }
    \label{fig:raw}
\end{figure*}
The data recorded by a camera sensor in RAW format exhibits a fundamentally linear relationship with the light intensity of the actual scene. This linearity is a core advantage in computational photography, particularly for HDR image synthesis. Leveraging this linear property of RAW data allows us to directly synthesize an HDR image by simply performing a linear combination of multiple exposures. This method circumvents the complex process of calculating and inverting the Camera Response Function (CRF). It significantly simplifies the HDR generation pipeline while improving fidelity.

In contrast to the RAW format, the conventional JPG format has inherent limitations for HDR applications. JPG files are typically stored and compressed using an 8-bit color depth, a process that results in a significant loss of color and luminance information. This loss of information is particularly severe under extreme lighting conditions, as illustrated in Figure~\ref{fig:raw}. The key differences are:

\begin{enumerate}
    \item \textbf{Shadow Detail and Color Fidelity:} In low-light environments, the RAW format, with its high bit depth (typically 12 or 14 bits), captures extensive detail in the dark regions. By increasing the exposure in post-processing, the original information can be recovered with minimal loss. Conversely, since this information is already discarded during the in-camera processing of a JPG image, attempting to boost its exposure does not recover meaningful detail and instead leads to severe color distortion, banding, and noise, as shown in Fig. \ref{fig:raw}.
    
    \item \textbf{Highlight Information Retention:} In highlight regions, while both a RAW-based HDR image and a JPG image may appear as pure white on a Standard Dynamic Range (SDR) display due to exceeding the display's maximum brightness, the amount of information they contain is fundamentally different. The RAW data fully retains the color and tonal information within these bright areas. When the exposure is reduced in post-processing, the details in these ``clipped" areas can be clearly recovered. As shown in the cloud section of Fig. \ref{fig:raw}, at an exposure compensation of -2EV, the RAW-based HDR image reveals subtle gradations in the clouds. The corresponding area in the JPG image, however, remains a flat white expanse because the information was permanently lost.
\end{enumerate}

RAW-based HDR synthesis offers unparalleled advantages over the JPG format, both in the recovery of shadow detail and the preservation of highlight information.

\section{Justification of Data Acquisition Setup}
\label{sec:supp_justification}

In this section, we provide a rigorous justification for our specific data acquisition protocol. We address the necessity of employing a dual-camera system over a single panoramic camera, the critical importance of spatial alignment, and the validity of our temporal swapping strategy.

\subsection{Necessity of the Dual-Camera System}
One might ask why we do not simply utilize a single high-resolution 360$^\circ$ camera and project the captures to perspective views to obtain scene images. While theoretically feasible, relying solely on projected 360$^\circ$ captures fails to meet the high-fidelity standards required for a photorealistic relighting benchmark for three primary reasons:

\begin{enumerate}
    \item \textbf{Effective Resolution:} Even with an 8K 360$^\circ$ capture, projecting the image to a standard 40mm field-of-view (FOV) yields an effective resolution significantly lower than that of the 24MP+ full-frame Sony A7 used in our rig. This loss of high-frequency detail would severely compromise the evaluation of texture preservation and generation.
    \item \textbf{Image Quality \& Dynamic Range:} Panoramic cameras typically utilize smaller sensors that introduce noise and chromatic aberration. More critically, they lack the dynamic range of the Sony A7's 14-bit RAW optical path. High dynamic range is essential for outdoor relighting tasks to accurately recover information in deep shadows and bright highlights.
    \item \textbf{Optical Artifacts:} 360$^\circ$ cameras rely on heavy distortion correction and stitching algorithms, which introduce resampling artifacts. Using a dedicated rectilinear lens ensures the benchmark data is free from such algorithmic interference.
\end{enumerate}

\begin{figure}[!t]
    \centering
    \includegraphics[width=0.7\linewidth]{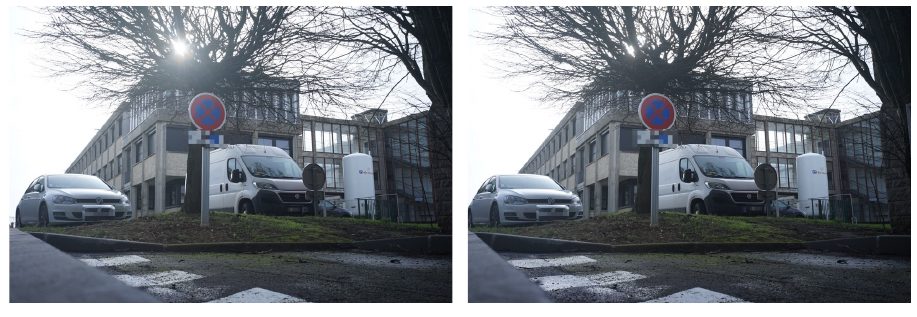}
    \caption{Side-by-side camera setup, a non-aligned envmap camera will record direct sun light, but scene camera records a shadow.}
    \label{fig:Alignment}
\end{figure}

\subsection{Necessity of Strict Spatial Alignment}
Precise co-location of the environment map camera (Insta360) and the scene camera (Sony A7) is a prerequisite for pixel-aligned evaluation. 

\paragraph{Spatial Parallax is Fatal.} 
A non-confocal setup (e.g., a standard side-by-side placement with a 10cm baseline) fundamentally alters occlusion relationships between the scene geometry and the light source. For instance, foliage that occludes the sun in the environment camera's view may not occlude it in the scene camera's view. This discrepancy creates ``false'' shadows in the Ground Truth, shadows that exist in the illumination map but not in the photograph (or vice versa). Such geometric inconsistencies render strict, pixel-aligned quantitative evaluation impossible.

\section{Methodology for Determining the Nodal Point (No-Parallax Point)}
\begin{figure*}
    \centering
    \includegraphics[width=\linewidth]{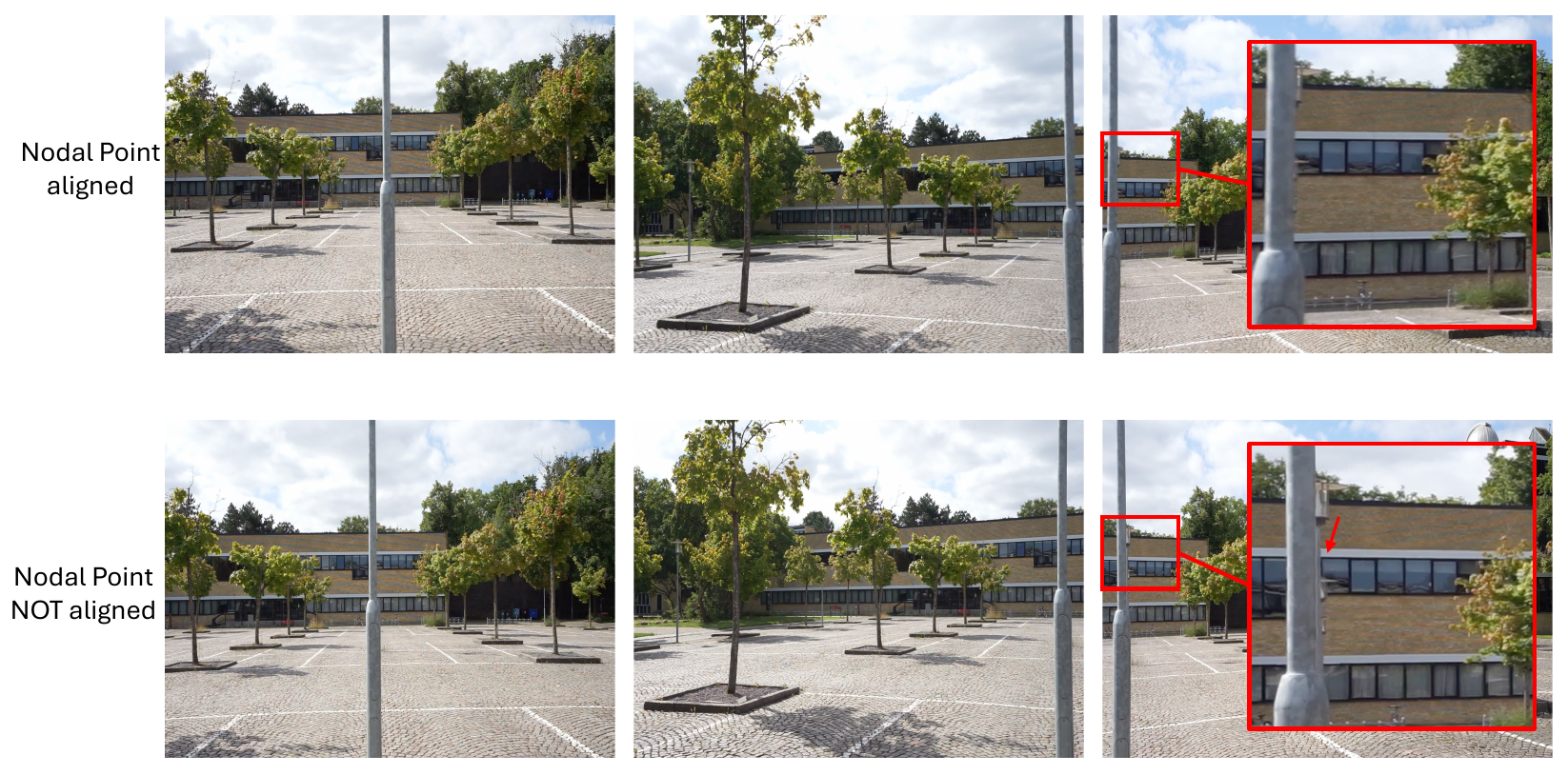}
    \caption{Nodal point alignment. When the camera is not positioned at the nodal point, rotating the camera causes the nearby utility pole to fail to occlude the poles behind it. In contrast, when the camera is at the nodal point, the nearby utility pole can occlude the poles behind it.}
    \label{fig:nodal_point}
\end{figure*}

In the domain of photographic composition, the elimination of parallax error is of paramount importance. Parallax manifests as an apparent displacement of foreground objects relative to the background as the camera is rotated. To mitigate this artifact, it is crucial to rotate the camera system around a specific pivot point. While colloquially referred to as the ``nodal point," the technically precise term for this locus is the center of the entrance pupil. The entrance pupil represents the virtual image of the physical aperture stop as viewed through the front elements of the lens. It is the conjugate point through which all chief rays appear to pass before refraction. Rotating the camera and lens assembly around the center of the entrance pupil ensures that the perspective remains consistent across multiple exposures, thereby facilitating seamless image stitching.

The empirical determination of the entrance pupil's location, or the no-parallax point, is a foundational procedure in panoramic photography. The following methodology outlines a systematic approach to identifying this point.

\subsection{Experimental Setup}

The camera must be securely mounted on a panoramic head affixed to a stable tripod. This specialized head allows for the adjustment of the camera's position along the longitudinal axis of the lens. For the purpose of this procedure, a focal length of 40mm was selected. Two distinct, vertically-oriented objects, positioned at different distances from the camera, were chosen as reference points. A lamppost and a more distant utility pole served as suitable subjects. The camera's position was initially adjusted so that, when viewed through the camera's live-view display, the nearer reference object precisely occluded the more distant one at the center of the frame.

\subsection{Procedure for Parallax Elimination}

The core of the methodology lies in an iterative process of rotation and observation. With the reference objects aligned, the camera is panned horizontally by an angle of approximately 30 degrees to the left and then to the right. The relative position of the two reference objects is carefully observed during this rotation. If the nearer object appears to shift its position relative to the farther object, parallax is present. This indicates that the axis of rotation is not coincident with the entrance pupil. Adjustments must then be made to the fore-aft position of the camera on the panoramic head, and the rotational test is repeated.

The objective is to achieve a state where, upon panning the camera to the left and right, the two reference objects remain in perfect alignment, with no discernible relative displacement, as shown in Fig. \ref{fig:nodal_point}. When this condition is met, the camera's axis of rotation is correctly aligned with the no-parallax point of the lens at the selected focal length. This position ensures that images captured from different angles will be free of parallax-induced stitching errors.

\begin{figure*}[!htb]
    \centering
    \includegraphics[width=0.9\linewidth]{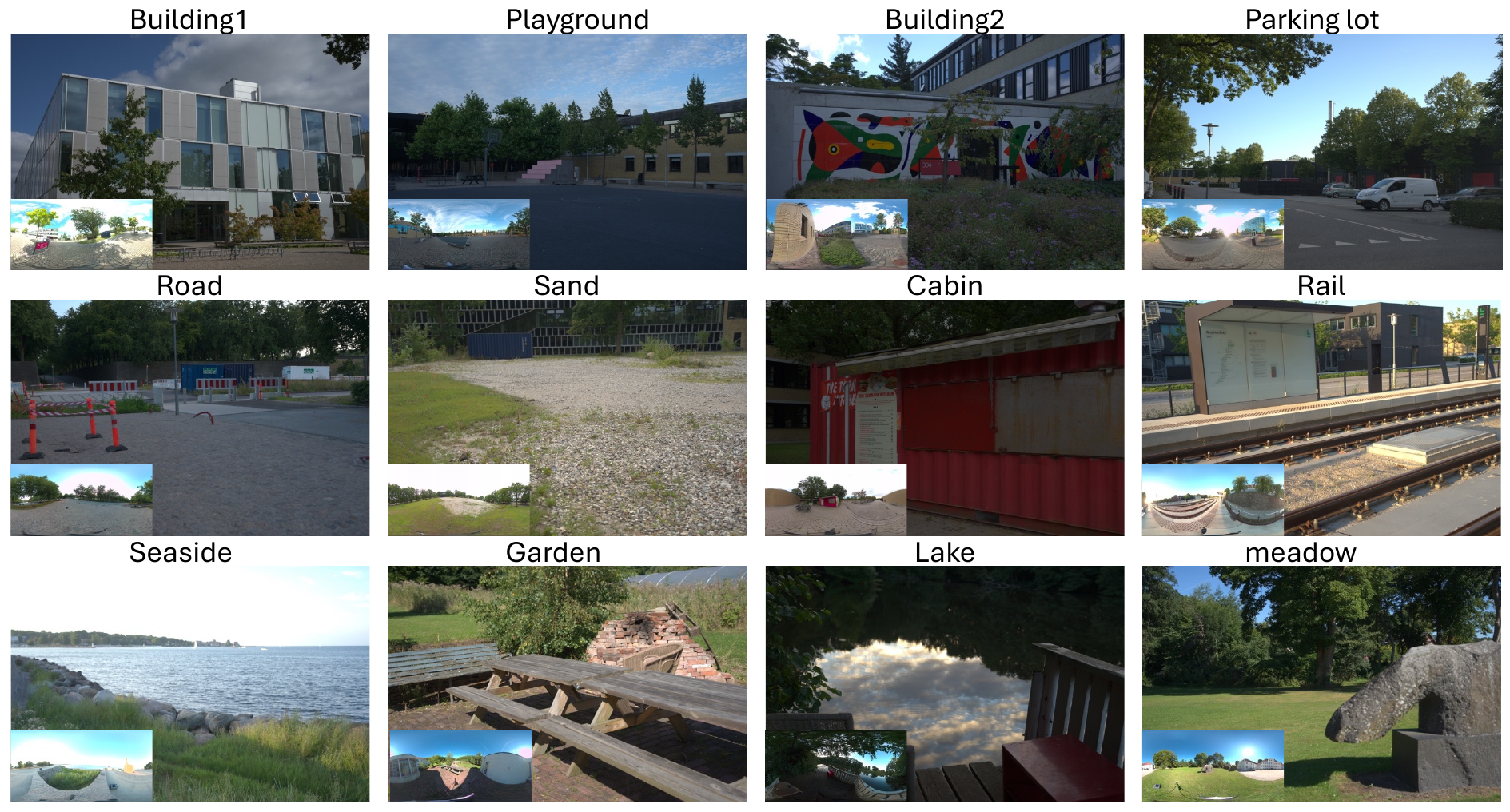}
    \caption{More examples from our dataset}
    \label{fig:showcase2}
\end{figure*}

\begin{figure}
    \centering
    \includegraphics[width=0.6\linewidth]{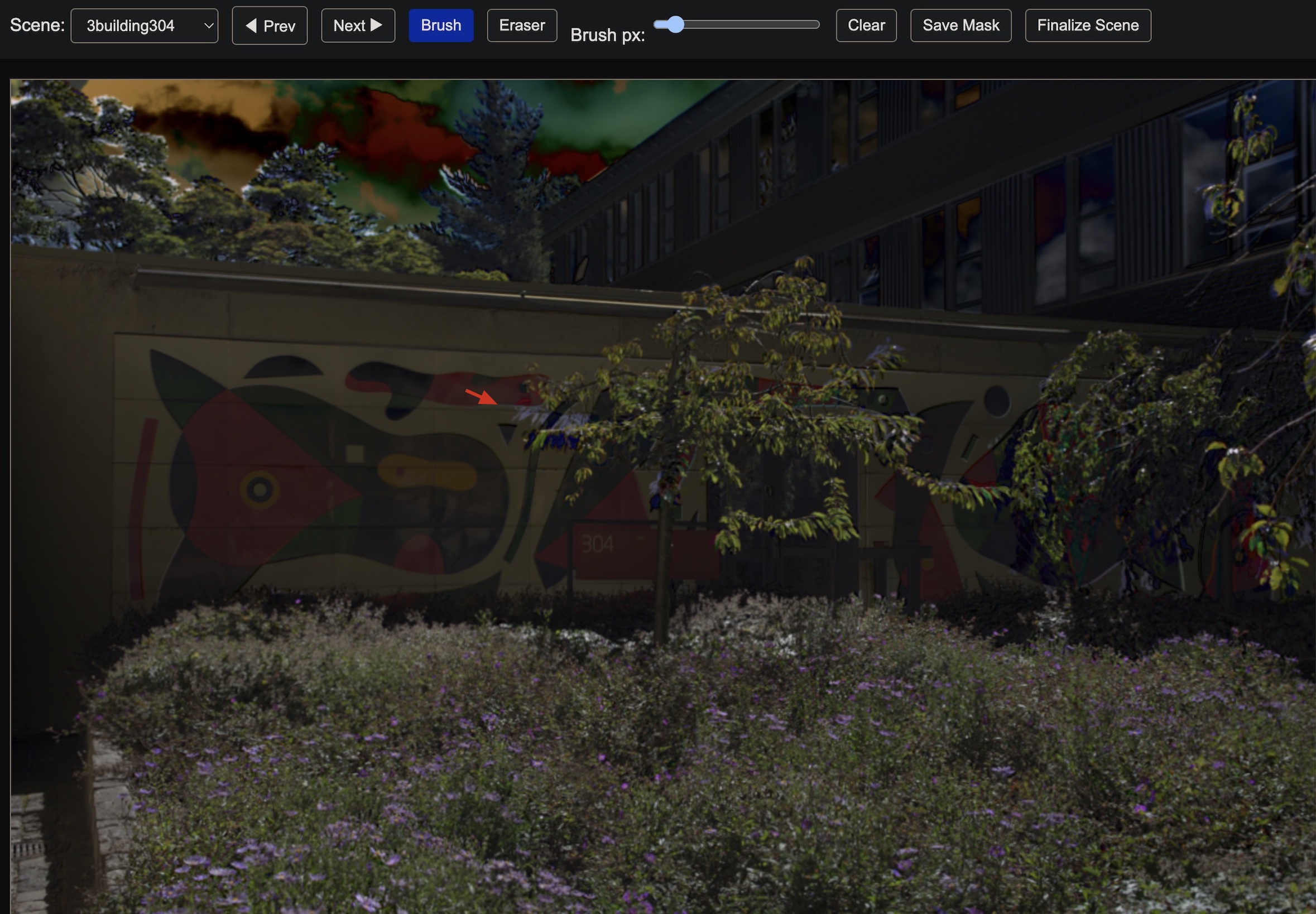}
    \caption{UI Interface for Marking Dynamic Scene Elements. Volunteers utilize a brush tool to create masks or an eraser tool to remove incorrectly marked regions. In the illustrated example, the areas indicated by red arrows correspond to discrepancies between the two photographs, highlighting regions where inconsistencies exist.}
    \label{fig:maskui2}
\end{figure}

\subsection{Details of Dynamic Scene Elements Annotation}
\label{sec:dynamic_masking}

A significant challenge in capturing longitudinal, ``in-the-wild" datasets is the presence of dynamic scene elements. While our capture rig ensures a static viewpoint, the long temporal intervals between acquisitions mean that elements such as wind blown foliage, grass, and cloud formations inevitably move. 

Although computational methods exist for image alignment (e.g., optical flow warping), applying such modifications would alter the pixel values and compromise the photometric integrity of our ground truth (GT) images. To preserve the dataset as a true GT reference, we opted instead to provide binary masks that identify these non-static regions. This approach allows researchers to optionally exclude these dynamic areas during metric computation, thereby isolating relighting performance from inconsistencies caused by scene motion.

We initially explored automated segmentation methods, including optical flow and contour detection, to identify these moving regions. However, these approaches produced unsatisfactory results, struggling with the subtle motions and complex natural textures present in our scenes. Consequently, we adopted a meticulous manual annotation process.

Our annotation pipeline is as follows:
\begin{enumerate}
    \item \textbf{Pairwise Comparison:} For each scene, annotators performed a sequential, pairwise comparison of adjacent time steps (e.g., $t_0$ vs. $t_1$, $t_1$ vs. $t_2$, etc.).
    \item \textbf{Difference Visualization:} To aid the human annotators, we generated absolute pixel-difference images for each pair. This visualization technique effectively accentuates the contours of misaligned objects, where pixel gradients are highest, making the boundaries of dynamic elements more conspicuous.
    \item \textbf{Manual Annotation:} Annotators manually painted masks over all identified dynamic regions for each image pair. The primary targets for masking were clouds and moving vegetation (leaves, branches, and grass).
    \item \textbf{Mask Aggregation:} The final mask for the entire scene is generated by computing the union of all pairwise masks. This ensures that any element that moved at any point during the capture sequence is included in the aggregate mask.
\end{enumerate}

We explicitly excluded two categories of dynamic effects from masking. First, water surfaces (e.g., lakes and seas) were not annotated due to their highly complex and stochastic textures, which are intractable to mask reliably. Second, dynamic reflections and refractions were intentionally left unmasked, as we consider the ability to model these complex, illumination dependent light transport phenomena to be a core aspect of the relighting challenge itself.
To facilitate the annotation of standard masks, we employ a dedicated user interface (UI) that enables manual annotation tasks (Fig. \ref{fig:maskui2}). The interface provides interactive tools for adjusting the brush size to generate masks, as well as an eraser function to correct erroneous regions.

\section{Differentiable Cook--Torrance Renderer}
\label{sec:supp_cook_torrance}

To evaluate the physics consistency of predicted G-buffers, we employ a fully differentiable Cook--Torrance microfacet model with split-sum approximation~\cite{cook1982reflectance}. Let the per-pixel surface properties be defined by basecolor $\mathbf{c}_b$, normal $\mathbf{n}$, roughness $\alpha$, and metallicity $m$, and let the environment illumination be given as a HDR map $L_\text{env}(\omega_i)$. The outgoing radiance along the view direction $\mathbf{v}$ is

\begin{equation}
L_o(\mathbf{v}) = L_\text{diff} + L_\text{spec},
\end{equation}

where the diffuse component is modulated by energy-conserving Lambertian reflection:

\begin{equation}
L_\text{diff} = (\mathbf{1} - m) \, (1 - F(\mathbf{n}\cdot\mathbf{v})) \, \mathbf{c}_b \, E_\text{diff}(\mathbf{n}),
\end{equation}

with $E_\text{diff}$ the diffuse irradiance obtained via spherical harmonics projection of the environment map. 

The specular component is

\begin{equation}
L_\text{spec} = F(\mathbf{n}\cdot\mathbf{v}) \, E_\text{spec}(\mathbf{n}, \mathbf{v}, \alpha),
\end{equation}

where $F$ is the Schlick Fresnel term

\begin{equation}
F(\mathbf{n}\cdot\mathbf{v}) = F_0 + (1-F_0) \, (1-\mathbf{n}\cdot\mathbf{v})^5, \quad 
F_0 = \text{lerp}(0.04, \mathbf{c}_b, m),
\end{equation}

and $E_\text{spec}$ is the prefiltered specular irradiance obtained by sampling the environment map along the reflection direction $\mathbf{r} = 2(\mathbf{n}\cdot\mathbf{v})\mathbf{n} - \mathbf{v}$ with roughness-dependent mipmapping. Energy conservation is enforced by weighting diffuse and specular components according to $F$ and $m$. The final output radiance is clamped to non-negative values for HDR consistency.

All operations are implemented in a differentiable manner, enabling backpropagation of reconstruction gradients from the rendered image $L_o$ to the predicted G-buffers during Diffusion Posterior Sampling (DPS) as described in Sec.~\ref{sec:method_dps}.

\end{document}